\documentclass[10pt,letterpaper]{article}
\usepackage[top=0.85in,left=2.75in,footskip=0.75in]{geometry}

\usepackage{amsmath,amssymb}

\usepackage{changepage}

\usepackage[utf8x]{inputenc}

\usepackage{textcomp,marvosym}

\usepackage{cite}

\usepackage{nameref,hyperref}

\usepackage{microtype}
\DisableLigatures[f]{encoding = *, family = * }

\usepackage[table]{xcolor}

\usepackage{array}

\usepackage{algorithm}
\usepackage[noend]{algpseudocode}
\usepackage{array}
\newcolumntype{P}[1]{>{\centering\arraybackslash}p{#1}}
\usepackage{multirow}

\newcolumntype{+}{!{\vrule width 2pt}}

\newlength\savedwidth

\raggedright
\usepackage[aboveskip=1pt,labelfont=bf,labelsep=period,justification=raggedright,singlelinecheck=off]{caption}

\makeatletter
\renewcommand{\@biblabel}[1]{\quad#1.}
\makeatother

\usepackage{lastpage,fancyhdr,graphicx}
\usepackage{epstopdf}
\fancyheadoffset[L]{2.25in}
\fancyfootoffset[L]{2.25in}
\begin{document}
\vspace*{0.2in}
\begin{flushleft}
{\Large
\textbf\newline{LoAdaBoost: loss-based AdaBoost federated machine learning with reduced computational complexity on IID and non-IID intensive care data} 
}
\newline
\\
Li Huang\textsuperscript{1,2},
Yifeng Yin\textsuperscript{3},
Zeng Fu\textsuperscript{4},
Shifa Zhang\textsuperscript{5,6},
Hao Deng\textsuperscript{7},
Dianbo Liu\textsuperscript{6,8*},
\\
\bigskip
\textbf{1} Academy of Arts and Design, Tsinghua University, Beijing, China
\\
\textbf{2} The Future Laboratory, Tsinghua University, Beijing, China
\\
\textbf{3} University of Huddersfield, Huddersfield, UK
\\
\textbf{4} University of California San Diego, San Diego, USA
\\
\textbf{5} Northeastern University, Boston, USA
\\
\textbf{6} Computer Science and Artificial Intelligence Laboratory,Massachusetts Institute of Technology, Cambridge, USA
\\
\textbf{7} Department of Anesthesia, Massachusetts General Hospital, Boston, USA
\\
\textbf{8} Harvard Medical School and Boston Children's Hospital, Boston, USA
\\

\bigskip

* dianbo@mit.edu

\end{flushleft}
\section*{Abstract}
Intensive care data are valuable for improvement of health care, policy making and many other purposes. Vast amount of such data are stored in different locations, on many different devices and in different data silos. Sharing data among different sources is a big challenge due to regulatory, operational and security reasons. One potential solution is federated machine learning, which is a method that sends machine learning algorithms simultaneously to all data sources, trains models in each source and aggregates the learned models. This strategy allows utilization of valuable data without moving them. One challenge in applying federated machine learning is the possibly different distributions of data from diverse sources. To tackle this problem, we proposed an adaptive boosting method named $LoAdaBoost$ that increases the efficiency of federated machine learning. Using intensive care unit data from hospitals, we investigated the performance of learning in IID and non-IID data distribution scenarios, and showed that the proposed $LoAdaBoost$ method achieved higher predictive accuracy with lower computational complexity than the baseline method.


\section*{Introduction}
Health data from intensive care units can be used by medical practitioners to provide health care and by researchers to build machine learning models to improve clinical services and make health predictions. But such data is mostly stored distributively on mobile devices or in different hospitals because of its large volume and high privacy, implying that traditional learning approaches on centralized data may not be viable. Therefore, federated learning that avoids data collection and central storage becomes necessary and up to now significant progress has been made.

In 2005, Rehak \textit{et al}. \cite{rehak2005model} established CORDRA, a framework that provided standards for an interoperable repository infrastructure where data repositories were clustered into community federations and their data were retrieved by a global federation using the metadata of each community federation. In 2011, Barcelos \textit{et al}. \cite{barcelos2011agent} created an agent-based federated catalog of learning objects (AgCAT system) to facilitate assess of distributed educational resources. Although little machine learning was involved in these two models, their practice of distributed data management and retrieval served as a reference for the development of federated learning algorithms. 

In 2012, Balcan \textit{et al}. \cite{balcan2012distributed} implemented probably approximately correct (PAC) learning in a federated manner and reported the upper and lower bounds on the amount of communication required to obtain desirable learning outcomes. In 2013, Richtárik \textit{et al}. \cite{richtarik2013distributed} proposed a distributed coordinate descent method named HYbriD for solving loss minimization problems with big data. Their work provided the bounds of communication rounds needed for convergence and presented experimental results with the LASSO algorithm on 3TB data. In 2014, Fercoq \textit{et al}. \cite{fercoq2014fast} designed an efficient distributed randomized coordinate descent method for minimizing regularized non-strongly convex loss functions and demonstrated that their method was extendable to a LASSO optimization problem with 50 billion variables. In 2015, Konecny \textit{et al}. \cite{konevcny2015federated} introduced a federated optimization algorithm suitable for training massively distributed, non-identically independently distributed (non-IID) and unbalanced datasets.

In 2016, McMahan \textit{et al}. \cite{mcmahan2016communication} developed the $FederatedAveraging$ ($FedAvg$) algorithm that fitted a global model with the training data left locally on distributed devices (known as clients). The method started by initializing the weight of neural network model at a central server, then distributed the weight to clients for training local models, and stopped after a certain number of iterations (also known as global rounds). At one global round, data held on each client would be split into several batches according to the predefined batch size; each batch was passed as a whole to train the local model; and an epoch would be completed once every batch was used for learning. Typically, a client was trained for multiple epochs and sent the weight after local training to the sever, which would compute the average of weights from all clients and distribute it back to them. Experimental results showed that $FedAvg$ performed satisfactorily on both IID and non-IID data and was robust to various datasets. 

More recently, Konevcny \textit{et al}. \cite{konevcny2016federated} modified the global model update of $FedAvg$ in two ways, namely structured updates and sketched updates. The former meant that each client would send its weight in a pre-specified form of a low rank or sparse matrix, whereas the latter meant that the weight would be approximated or encoded in a compressed form before sending to the server. Either way aimed at reducing the uplink communication costs, and experiments indicated that the reduction can be two orders of magnitude.In addition, Bonawitz \textit{et al}. \cite{bonawitz2016practical} designed the \textit{Secure Aggregation} protocol to protect the privacy of each client's model gradient in federated learning, without sacrificing the communication efficiency. Later, Smith \textit{et al}. \cite{smith2017federated} devised a systems-aware optimization method named MOCHA that considered simultaneously the issues of high communication cost, stragglers, and fault tolerance in multi-task learning. Zhao \textit{et al}. \cite{zhao2018federated} addressed the non-IID data challenges in federated learning and presented an improved version of FedAvg with a data-sharing strategy whereby the test accuracy could be enhanced significantly with only a small portion of globally shared data among clients. The strategy required the server to prepare a small holdout dataset $G$ (sampled from IID distribution) and globally share a random portion $\alpha$ of $G$ with all clients. The size of $G$ was defined as $\beta=\frac{\text{number of examples in \textit{G}}}{\text{total number of examples in all clients}} \times 100\%$. There existed two trade-offs: first, test accuracy and $\alpha$; and second, test accuracy and $\beta$. A rule of thumb was that the larger $\alpha$ or $\beta$ was, the higher test accuracy would be achieved. It is worth mentioning that since $G$ was a separate dataset from the clients’ data, sharing it would not be a privacy breach. Since no specific name was given to this method in Zhao et al.’s literature \cite{zhao2018federated}, we referred to it as “FedAvg with data-sharing" in our study. Bagdasaryan \textit{et al}. \cite{bagdasaryan2018backdoor} designed a novel model-poisoning technique that used \textit{model replacement} to backdoor federated learning. Liu et al. used a federated transfer learning strategy to balance global and local learning\cite{liu2018fadl,huang2019patient,liu2019two,liu2018artificial,liu2019confederated}. 

Most of the previously published federated learning methods focused on optimization of a single issue such as test accuracy, privacy, security or communication efficiency; yet none of them considered the computation load on the clients. This study took into account three issues in federation learning, namely, the local client-side computation complexity, the communication cost, and the test accuracy. We developed an algorithm named Loss-based Adaptive Boosting $FederatedAveraging$ (\textit{LoAdaBoost FedAvg}), where the local models with a high cross-entropy loss were further optimized before model averaging on the server. To evaluate the predictive performance of our method, we extracted the data of critical care patients' drug usage and mortality from the Medical Information Mart for Intensive Care (MIMIC-III) database \cite{johnson2016mimic} and the eICU Collaborative Research Database \cite{pollard2018eicu}. The data were partitioned into IID and non-IID distributions. In the IID scenario \textit{LoAdaBoost FedAvg} was compared with $FedAvg$ by McMahan \textit{et al}. \cite{mcmahan2016communication}, while in the non-IID scenario our method was complemented by the data-sharing concept before being compared with FedAvg with data-sharing by Zhao \textit{et al}. \cite{zhao2018federated}. Our primary contributions include the application of federated learning to health data and the development of the straightforward \textit{LoAdaBoost FedAvg} algorithm that had better performance than the state-of-the-art $FedAvg$ approach.

\section*{Materials and methods}
\subsection*{FedAvg: the baseline in IID scenario}
\begin{figure}[!h]
\centering
\includegraphics[width=0.8\linewidth]{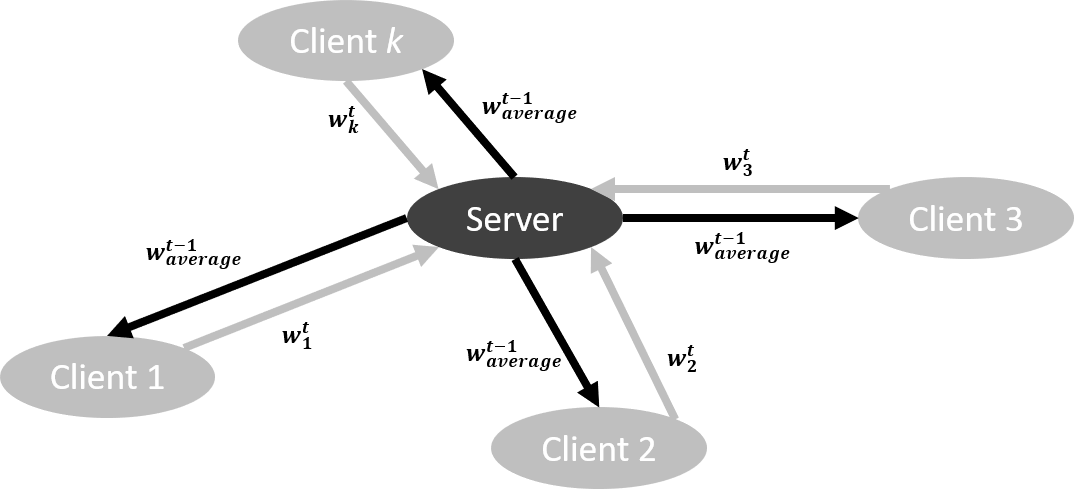}
\caption{\bf communication between the clients and the server under $FedAvg$.}
\label{fig:5}
\end{figure}

Developed by McMahan \textit{et al}. \cite{mcmahan2016communication}, the $FedAvg$ algorithm trained neural network models via local stochastic gradient descent (SGD) on each client and then averaged the weight of each client model on a server to produce a global model. This local-training-and-global-average process was carried out iteratively as follows. At the $t$\textsuperscript{th} iteration, a random $C$ fraction of the clients were selected for computation: the server first sent the average weights at the previous iteration (denoted $w_{average}^{t-1}$) to the selected clients (except for the $1$\textsuperscript{st} iteration where the clients started its model from the same random weight initialization); each client independently learnt a neural network model initialized with $w_{average}^{t-1}$ on its local data divided into $B$ minibatches for $E$ epochs, and then reported the learned weights (denoted $w_k^t$ where $k$ was the client index) to the server for averaging (see Figure \ref{fig:5}). The global model was updated by the average weights of each iteration. $FedAvg$ was utlized as the baseline method in IID scenario where both the training and test data were identically independently distributed.

\subsection*{FedAvg with data-sharing: the baseline in non-IID scenario}
\begin{figure}[!h]
\centering
\includegraphics[width=0.5\linewidth]{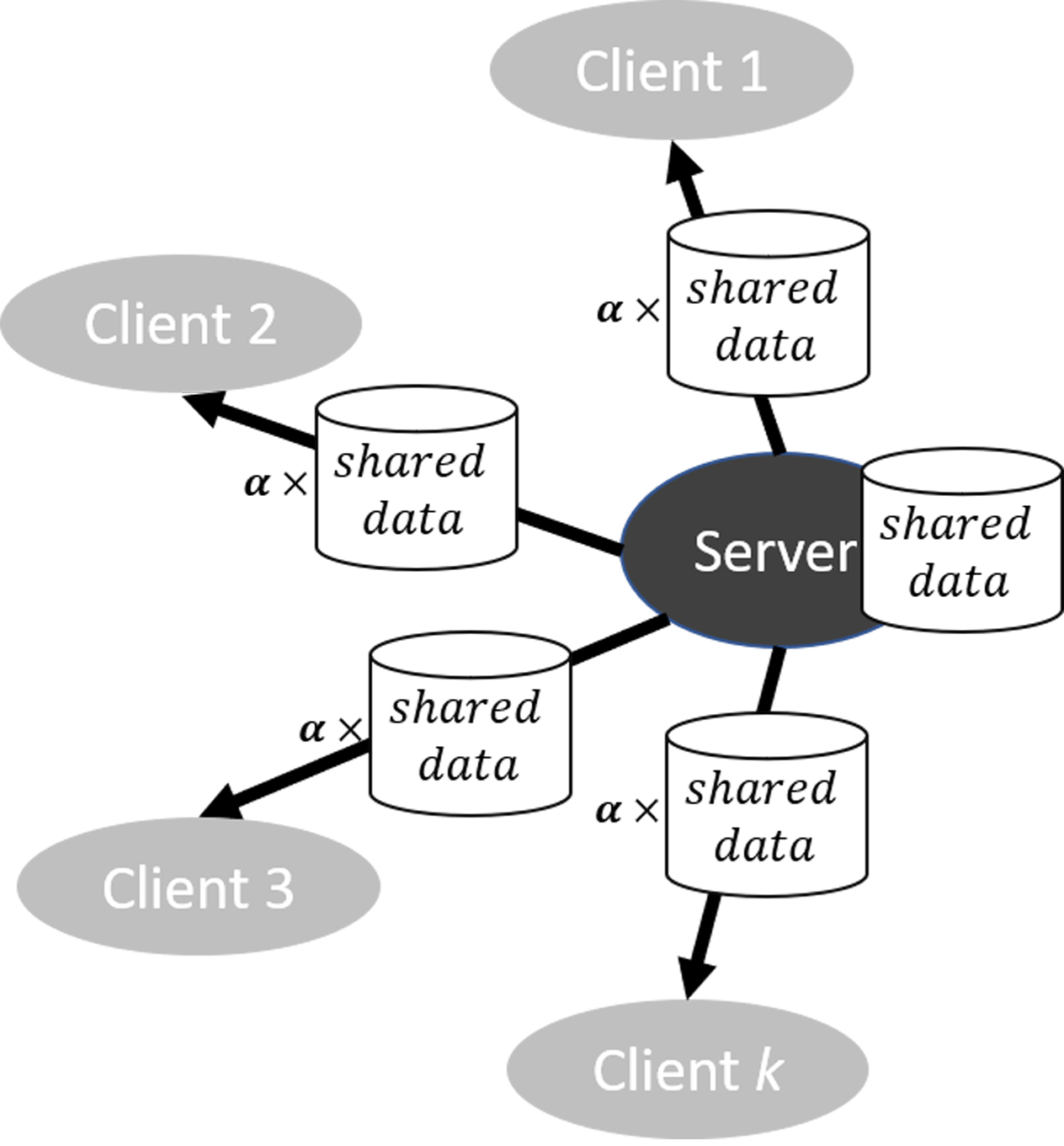}
\caption{\bf $FedAvg$ complemented by the data-sharing strategy: distribute shared data to the clients at initialization.}
\label{fig:6}
\end{figure}

As demonstrated in the literature\cite{mcmahan2016communication}, $FedAvg$ exhibited satisfactory performance with IID data, but its accuracy could drop substantially when trained on non-IID data. This was because, with non-IID sampling, stochastic gradient could no longer be regarded as an unbiased estimate of the full gradient according to Zhao \textit{et al}. \cite{zhao2018federated}. To address the challenge, they proposed an improved version of FedAvg: a data-sharing strategy complemented $FedAvg$ via globally sharing a small subset of training data between all the clients (see Figure \ref{fig:6}). Stored on the server, the shared data was a dataset distinct from the clients’ data and assigned to clients when $FedAvg$ was initialized. Thus, this strategy improved $FedAvg$ with no harm to privacy and little addition to the communication cost. The strategy had two parameters that were $\alpha$, the random fraction of the globally-shared data distributed to each client, and $\beta$, the ratio of the globally-shared data size to the total client data size. Raising these two parameters could lead to a better predictive accuracy but meanwhile make federated learning less decentralized, reflecting a trade-off between non-IID accuracy and centralization. In addition, it is worth mentioning that Zhao \textit{et al}. also introduced an alternative initialization for their data-sharing strategy: the server could train a warm-up model on the globally shared data and then distribute the model’s weights to the clients, rather than assigning them with the same random initial weights. In this work, we kept the original initialization method to leave all computation on the clients. $FedAvg$ with data-sharing was used as the baseline method in non-IID scenario where both the training and test data came from non-identically independently distributions.

\subsection*{LoAdaBoost FedAvg}
\begin{figure}[!h]
\centering
\includegraphics[width=0.8\linewidth]{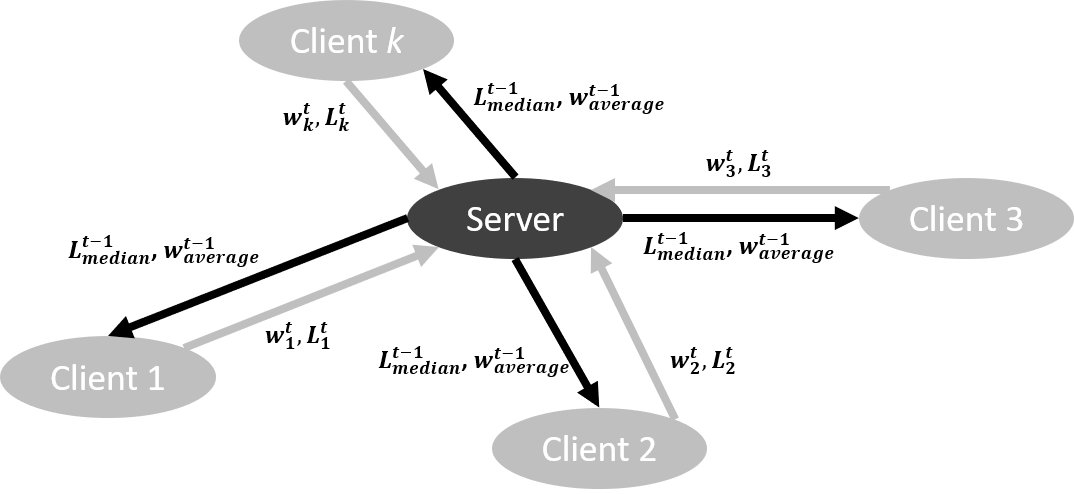}
\caption{\bf Communication between the clients and the server under $LoAdaBoost$ $FedAvg$.}
\label{fig:7}
\end{figure}

We devised a variant of $FedAvg$ named \textit{LoAdaBoost FedAvg} that was based on cross-entropy loss to adaptively boost the training process on those clients appearing to be weak learners. Since in our study the data labels were either 0 (survival) or 1 (expired), binary cross-entropy loss was adopted as the error measure of model-fitting and calculated as
\begin{equation}\label{eq:1}
    -\sum_{i=1}^{N} [y_i\log {f(x_i)}+(1-y_i)\log {(1-f(x_i))}]
\end{equation}
where $N$ was the total number of examples, $x_i$ was the input drug feature vector, $y$ was the binary mortality label, and $f$ was the federated learning model. The objective function of each client model under $FedAvg$ and $LoAdaBoost$ learning was to minimize Equation \ref{eq:1}, which measured goodness-of-fit: the lower the loss was, the better a model was fitted. Our method utilized the median cross-entropy loss $L_{median}^{t-1}$ of clients that participated in the previous global round $t-1$ as a criterion for boosting Client $k$. Retraining for more epochs would be incurred if, after training for $E/2$ epochs at the current global round $t$, Client $k$’s cross-entropy loss $L_k^{t,0}$ was above $L_{median}^{t-1}$. The reason for using the median loss rather than average lied in that the latter was less robust to outliers that were significantly underfitted or overfitted client models. Communication between clients and the server under $LoAdaBoost$ is demonstrated in Figure \ref{fig:7}. Not only the model weights but also the cross-entropy losses were communicated between the clients and the server. At the $t$\textsuperscript{th} iteration, the server delivered the average weights $w_{average}^{t-1}$ and the median loss $L_{median}^{t-1}$ obtained at the $t-1$\textsuperscript{th} iteration to each client; then, each client learnt a neural network model in a loss-based adaptive boosting manner, and reported the learnt weights $w_k^{t,r}$ and the cross-entropy loss $L_k^{t,r}$ to the server. The global model was parametrized by the average of $w_k^{t,r}$. 

Algorithm \ref{alg:LoAdaBoost} shows how $LoAdaBoost$ worked in detail. The server started a neural network model by randomly initializing the weight $w_0$, which was then distributed to each client. The initial value of median training loss ($L_{median}^0$) of client models was set to 1.0, and the number of clients participating in federated learning ($m$) was determined by the product of the client percentage $C$ and the total client count $K$. At least one client model would be trained in each global round. At the $t$th round, Client $k$ was initialized with the average weight from the $t-1$th round $w_{average}^{t-1}$, and trained on the local data for $E/2$ epochs to obtain weight $w_k^{t,0}$ and loss $L_k^{t,0}$ before retraining. For odd $E$, $E/2$ would be rounded up to the nearest integer. If $L_k^{t,0}$ was not greater than the median loss from the previous round $L_{median}^{t-1}$, computation on Client $k$ would be finished, with $w_k^{t,0}$ and $L_k^{t,0}$ sent to the server. Otherwise, the client would be retrained for another $E/2$ epochs. Now, the new loss was denoted $L_k^{t,1}$ where the superscript $1$ indicated the first retraining round. If $L_k^{t,1}$ was still above $L_{median}^{t-1}$, Client $k$ would be retrained for $E/2-1$ more epochs. This process was repeated for retraining round $r$=1,2,3,..., each round for $\max$($E/2-r+1, 1$) epochs, and stopped until the retrained loss $L_k^{t,r}$ dropped below $L_{median}^{t-1}$ or the total number of epochs (including initial training and retraining) reached $3E/2$. Lastly, $L_k^{t,0}$ and the final $w_k^{t,r}$ were sent to the server.
\begin{algorithm}
\caption{{\fontfamily{qcr}\selectfont
LoAdaBoost FedAvg.} The $K$ clients are indexed by $k$, $C$ is the fraction of clients that perform computation at each global round, and $E$ is the number of local epochs}\label{alg:LoAdaBoost}
\begin{algorithmic}[1]
    \State server initializes  weight $w_0$
    \State $L_{median}^0\gets 1.0$
    \State $m\gets{\max(C\cdot K, 1)}$
    \For {each global round $t=1,2,...$}
        \State $S_t\gets$(random set of $m$ clients)
        \For {each client $k \in S$ \textbf{in parallel}}
            \State train neural network model $f_k$ for $\frac{E}{2}$ epochs to obtain $w_k^{t,0}$ and $L_k^{t,0}$
            \If{$L_k^t \leq L_{median}^{t-1}$}
                \State $w_k^t\gets w_k^{t,0}$
            \Else
                \State $w_k^t\gets$ Retrain($f_k$, $E$, $L_{median}^{t-1}$)
            \EndIf
            \State \textbf{return} $w_k^t$, $L_k^{t,0}$ to server
            \EndFor
        \EndFor
\State
\Function{Retrain}{$f_k$, $E$, $L_{median}^{t-1}$}
    \For {each retrain round $r=1,2,...$}
        \State train $f_k$ for $\max$($\frac{E}{2}-r+1, 1$) epochs to obtain $w_k^{t,r}$ and $L_k^{t,r}$
        \If{$L_k^{t,r}>L_{median}^{t-1}$ \textbf{or} total training epochs $>\frac{3E}{2}$}
            \State \textbf{return} $w_k^{t,r}$
        \EndIf
    \EndFor
\EndFunction
\end{algorithmic}
\end{algorithm}

Depending on its cross-entropy loss, each client would be trained for at least $E/2$ epochs and at most $3E/2$ epochs. We set the maximum training epochs to $3E/2$ to control computational complexity of $LoAdaBoost$, aiming to prevent it from running more average epochs than $FedAvg$. The median cross-entropy loss of clients from the $t-1$th global round $L_{median}^{t-1}$ was used as the criterion for retraining clients at the $t$th round. In the worst-case scenario, no improvement of training loss was made on each client after the initial $E/2$ epochs, and about half of the clients were retrained for the full $E$ additional epochs. Thus, the expected number of epochs per client per global round would be at most $E$. 

$LoAdaBoost$ was adaptive in the sense that the performance of a poorly-fitted client model after the first $E/2$ epochs was boosted via continuous retraining for a decaying number of epochs. The quality of training was determined by comparing the model’s loss $L_k^{t,r}$ with the median loss $L_{median}^{t-1}$. In this way, our method was able to ensure that the losses of most (if not all) client models would be lower than the median loss at the prior iteration, thereby making the learning process more effective. In addition, because at one iteration only a few of the client models were expected to be trained for the full $3E/2$ epochs, the average number of epochs run on each client would be less than E, meaning a smaller local computational load under our method than that of $FedAvg$. Furthermore, since both $L_{median}^{t-1}$ and $L_k^{t,r}$ were a single value transferred at the same time with $w_k^{t,r}$ between the server and Client $k$, little additional communication cost would be incurred by our method.

Similar to other stochastic optimization-based machine learning methods \cite{zhao2018federated,bottou2010large,rakhlin2012making,ghadimi2013stochastic}, an important assumption for our approach to work satisfactorily was that the stochastic gradient on the clients’ local data was an unbiased estimate of the full gradient on the population data. This held true for IID data but broke for non-IID. In the latter case, an optimized client model with low losses did not necessarily generalize well to the population, implying that reducing the losses through adding more epochs to the clients was less likely to enhance the global model’s performance. This non-IID problem could be alleviated by combining \textit{LoAdaBoost FedAvg} with the data-sharing strategy, because the local data became less non-IID when integrated with even a small portion of IID data.

\subsection*{The MIMIC-III database}
The performance evaluation concerned with the MIMIC-III database \cite{johnson2016mimic}, which contains health information for critical care patients at a large tertiary care hospital in the US. Included in MIMIC-III are 26 tables of data ranging from patients’ admissions, to laboratory measurements, diagnostic codes, imaging reports, hospital length of stay and more. We processed three of these tables, namely ADMISSIONS, PATIENTS and PRESCRIPTIONS, to obtain two new tables as follows:
\begin{itemize}
\item 	ADMISSIONS and PATIENTS were inner-joined on $SUBJECT\_ID$ to form the PERSONAL\_INFORMATION table which recorded $AGE\_GROUP$, $GENDER$ and the survival status ($MORTALITY$) of all patients.
\item 	Each patient’s usage of DRUGS during the first 48 hours of stay (that is, $STARTDATE-ENDDATE$ = two days) at the hospital was extracted from PRESCRIPTIONS to give the SUBJECT\_DRUG\_TABLE table.
\end{itemize}
Further joining these two tables on $SUBJECT\_ID$ gave a dataset of 30,760 examples, from which we randomly selected 30,000 examples to form the evaluation dataset where $DRUGS$ were the predictors and $MORTALITY$ was the response variable. The summary of this dataset was provided in Table \ref{tab:6}.

\begin{table}[!ht]
\begin{adjustwidth}{-2.25in}{0in} 
\centering
\caption{\label{tab:6}{\bf Summary of the evaluation dataset.}}
\begin{tabular}{ |c|c|c| }
 \hline
 \empty & representation &  count\\
 \hline
$SUBJECT\_ID$ & integer: IDs ranging from 2 to 99,999 & 30,000 \\
 \hline
$GENDER$	& binary: 0 for female and 1 for male	& 17,284/12,716 \\
 \hline
$AGE\_GROUP$ &	binary: 0 for ages less than or equal to 65 and 1 for greater	& 13,947/16,053 \\
 \hline
$MORTALITY$	& binary: 0 for survival and 1 for expired	& 20,841/9,159 \\
 \hline
$DRUGS$	& binary: 0 for not prescribed to patients and 1 for prescribed	& 2814 dimensions \\
 \hline
\end{tabular}
\end{adjustwidth}
\end{table}

The drug feature contained 2814 different drugs prescribed to the patients. Table \ref{tab:7} shows the first six drugs D5W (that is, 5\% dextrose in water), Heparin Sodium, Nitro-glycerine, Docusate Sodium, Insulin and Atropine Sulphate. If a drug was prescribed to a patient (identified by $SUBJECT\_ID$), the corresponding cell in the table would be marked 1, and 0 otherwise. For instance, Patient 9 was given D5W and Insulin while none of the first six drugs were offered to Patient 10. 

\begin{table}[!ht]
\begin{adjustwidth}{-2.25in}{0in} 
\centering
\caption{\label{tab:7}\bf {Example rows and columns of DRUGS.}}
\begin{tabular}{ |c|c|c|c|c|c|c|c| }
 \hline
 SUBJECT$\_$ID & D5W &  Heparin Sodium & Nitro-glycerine & Docusate Sodium & Insulin & Atropine Sulphate & ...\\
 \hline
... & ...&...&...&...&...&...&... \\
 \hline
9 & 1 & 0 & 0 & 0 & 1 &	0 & ... \\
 \hline
10 & 0 & 0 & 0 & 0 & 1 &	0 & ... \\
 \hline
11 & 0 & 0 & 0 & 1 & 1 &	0 & ... \\
 \hline
12 & 1 & 0 & 0 & 0 & 1 &	0 & ... \\
 \hline
13 & 1 & 1 & 1 & 1 & 1 &	1 & ... \\
\hline
\end{tabular}
\end{adjustwidth}
\end{table}

The evaluation dataset was shuffled and split into a training set of 27,000 examples and a holdout set of 3,000 examples for implementing data-sharing strategy. As with the literature \cite{mcmahan2016communication}, the training set was partitioned over 90 clients in two ways: IID in which the data was randomly divided into 90 clients, each consisting of 300 examples; and non-IID in which the data was firstly sorted according to $AGE\_GROUP$ and $GENDER$, and then split into equal-sized 90 clients. Using the skewed non-IID data, we would be able to assess the robustness of our model to scenarios when IID data assumption cannot be made, which is more realistic in the healthcare industry. 

\subsection*{Parameter sets}
The neural network trained on each client consisted of three hidden layers with $20$, $10$ and $5$ units, respectively, using the rectified linear unit (ReLu) activation functions. There were $56,571$ parameters in total. The stochastic optimizer chosen in this study was Adaptive Moment Estimation (Adam), which requires less memory and is more computationally efficient according to empirical results \cite{kingma2014adam}. We used the default parameter set for Adam in the Keras framework: the learning rate $\eta=0.001$ and the exponential decay rates for the moment estimates $\beta_1=0.9$ and $\beta_2=0.999$. In addition, while setting the minibatch size $B$ to 30, we experimented with the number of epochs $E=5$,$10$ and $15$ and the fraction of clients $C$=10\%, 20\%, 50\% and 100\% (same as in the work of McMahan \textit{et al}. \cite{mcmahan2016communication}).

As for parameters of the data-sharing strategy, we experimented with various combinations of $\alpha$s (10\%, 20\% and 30\%) and $\beta$s (1\%, 2\% and 3\%). For instance, $\alpha$=10\% and $\beta$=1\% meant only $0.1\%$ (that is, 270 examples) of the total non-IID data were shared across the clients, each receiving 27 random examples. Small $\alpha$ and $\beta$ were chosen to implement the data-sharing strategy because we only sought to demonstrate that data-sharing could narrow the performance gap between learning on IID and non-IID data. Large values were unnecessary for this purpose, though both $\alpha$ and $\beta$ could be increased to further enhance the performance, at the expense of decentralization \cite{zhao2018federated}.

\subsection*{Evaluation metrics}
Evaluation metrics were twofold. First, the area under the ROC curve (AUC) was used to assess the predictive performance of a federated learning model. Here, ROC stands for the receiver operating characteristic curve, a plot of the true positive rate (TPR) against the false positive rate (FPR) at various thresholds. For a given threshold, TPR was the ratio of the number of mortalities predicted by the global model to the total number of mortalities in the test dataset, while FPR was calculated as $1-specificity$ where $specificity$ was the ratio of the number of predicted survivals to the total number of survivals. In our study, 10-fold cross validation was performed to reduce the level of randomness. In IID evaluation, we partitioned the MIMIC III data of 27,000 examples into 90 clients (each holding 300 examples) and further randomly split the clients into 10 folds (each containing 9 clients). In non-IID evaluation, the data was sorted by patients' age and gender before partitioning. Then, each fold was regarded as the test data in turn and the remaining nine folds were used to train $FedAvg$ and $LoAdaboost$. Predictions for every fold were recorded and compared against the true labels, and AUC ROC at convergence was calculated. This process was repeated for five times, resulting in a set of five cross-validation AUC values. $FedAvg$ and $LoAdaboost$ were compared in terms of average and standard deviation of these values.

Second, we defined average epochs of clients as the expected number of epochs to run on a single client in a complete federated learning process and used the metric to measure the computational complexity of federated learning algorithms.
\begin{equation}
    \frac{\sum_t^T \sum_k^m (\frac{E}{2}+\text{retraining epochs for Client $k$ at \textit{t}th global round})}{m}
\end{equation}
where $T$ was the total number of global rounds taken by an algorithm to converge and $m$ was the number of clients participating in computation at each global round. Under $FedAvg$, average epochs would be a constant value of $E$ times the number of global rounds, while under our adaptive method it would be varying because each client expectedly ran for a different number of epochs. In the experiments, we set a maximum number of global rounds, then carried out 10-fold cross validation with different random seeds for five times, and finally calculated cross-validation AUCs and average epochs.

\section*{Results}
\begin{figure}[!h]
\centering
\includegraphics[width=0.8\linewidth]{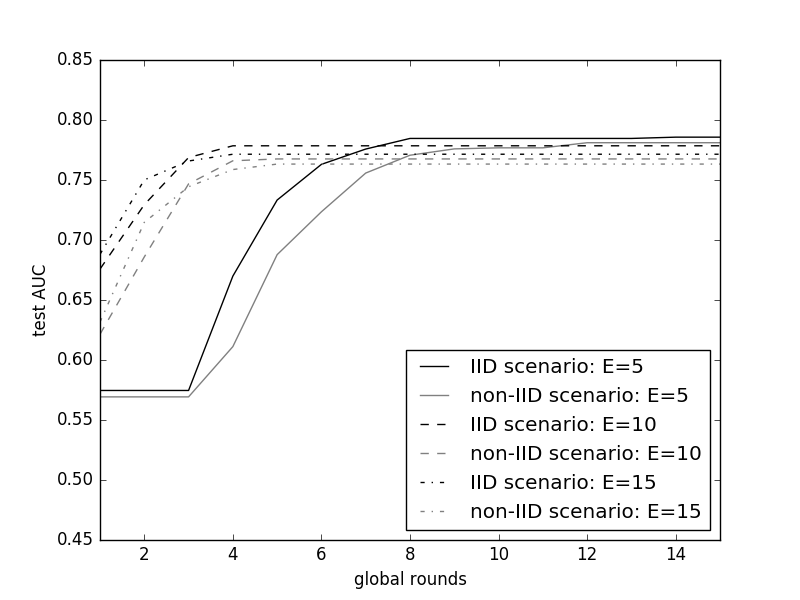}
\caption{\bf Performance gap between IID and non-IID data.}
\label{fig:1}
\end{figure}

\textit{LoAdaBoost} was evaluated against the baseline $FedAvg$ algorithm in IID scenario and $FedAvg$ with data-sharing in non-IID sceniaro. We adpoted the data-sharing strategy on non-IID data because there was a performance gap between the two scenarios, as depicted in Figure \ref{fig:1}. The figure shows test AUCs versus global rounds during a single cross-validation run of FedAvg with varying numbers of local epochs $E$. Same as the work by McMahan \textit{et al}.\cite{mcmahan2016communication}, each curve in the figure was made monotonically increasing via taking the highest test-set AUC achieved over all previous global rounds. It is apparent that $FedAvg$ on IID data consistently exhibited a higher test AUC than on non-IID data for all different $E$s.

Throughout the evaluation, 10-fold cross-validation with five repetitions was carried out to obtain an accurate estimate of predictive performance: 27,000 examples of the MIMIC III data were divided into 90 equally-sized clients, which were further randomly split into 10 folds, each containing nine clients. In cross validation, each fold was regarded as the test set in turn and the other nine folds were used to train models. The remaining 3,000 examples were utilized as the holdout set to implement the data-sharing strategy in non-IID scenario.

\subsection*{Evaluation in IID scenario}
\begin{figure}[!h]
\centering
\includegraphics[width=0.8\linewidth]{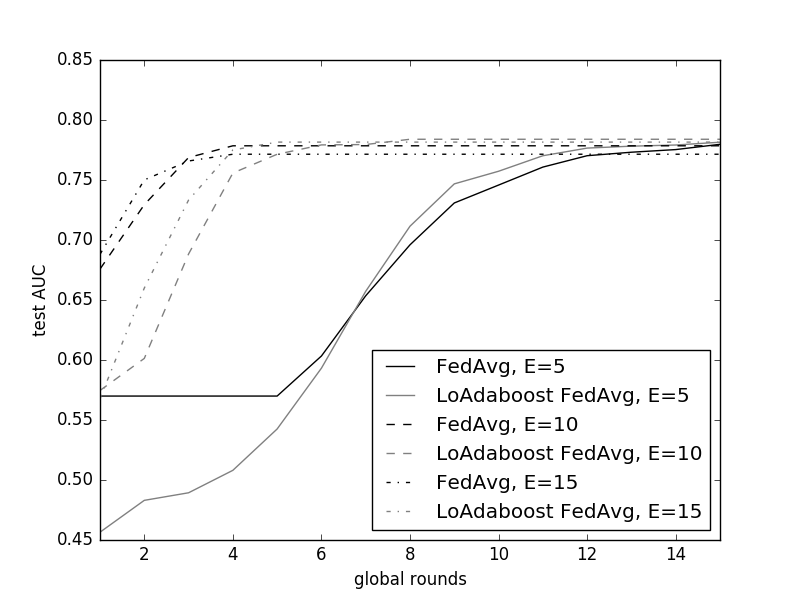}
\caption{{\bf Comparison of $FedAvg$ and $LoAdaboost$ on IID data.} $LoAdaBoost$ converged slightly slower than $FedAvg$, but to a higher test AUC.}
\label{fig:2}
\end{figure}

Figure \ref{fig:2} compares the predictive performance (test AUC versus global rounds) of $FedAvg$ and $LoAdaboost$ with $C$=10\% and $E$=5, 10 and 15 using the same training and test data as in Figure \ref{fig:1}. Given the same $E$, our method seemed to converge slightly slower (lagging a couple of global rounds) but nonetheless to a higher test AUC than $FedAvg$.

We speculate the reason for this lagged convergence as follows. At the first few global rounds where each client model was underfitting, learning $FedAvg$ would be more efficient because each client was trained to the full five epochs. After a few global rounds, some client models would start to be overfitted and impose an adverse effect on the predictive performance of the averaged model on the server. So, learning speed of $FedAvg$ would be lowered. On the other hand, our method would be less affected by individual overfitted client models, because the loss-based adaptive boosting mechanism would enable underfitted models to be trained for more epochs and overfitted ones to be trained for less epochs than five. Finally, when all clients became overfitted, FedAvg and our method would cease to learn, though the convergence AUC for the latter would be higher.

In addition, both algorithms converged faster with a larger value of $E$. With $E$ equal to 5, they began to converge at the 15th global round; with E equal to 10, they had already converged at the 10th round; and with $E$ equal to 15, at the 5th round $FedAvg$ had already converged while our method began to converge to a higher point. 

To make the superiority of our method more credible, 10-fold cross validation was carried out with different combinations of $C$ and $E$, and was repeated for five times under each experimental setting. Wilcox signed rank test was performed on the AUC sets for FedAvg and our method. Average cross validation AUC (with standard deviation), average epochs, and p-values for the statistical test are shown in Table \ref{tab:1}.
\begin{table}[!ht]
\begin{adjustwidth}{-2.25in}{0in} 
\centering
\caption{\label{tab:1}\bf IID scenario: 10-fold cross validation results with varying $C$ and $E$.}
\begin{tabular}{|P{1cm}|P{0.7cm}|P{3.0cm}|P{2.3cm}|P{3.0cm}|P{2.3cm}|P{1cm}|}
\hline
\multirow{2}{*}{$C$}&\multirow{2}{*}{$E$}&\multicolumn{2}{c|}{FedAvg}&\multicolumn{2}{c|}{LoAdaBoost}&\multirow{2}{*}{$p$-value} \\
\cline{3-6}
& & AUC & average epochs & AUC & average epochs & \\
\hline
\multirow{3}{*}{10\%} & 5 & 0.7891+-0.0002 & 75 & 0.7940+-0.0001 & 68 & 0.03\\
\cline{2-7}
&10& 0.7876+-0.0010 & 100 & 0.7900+-0.0007 & 73 & 0.03 \\
\cline{2-7}
&15& 0.7897+-0.0006 & 75 & 0.7907+-0.0010 & 52 & 0.03\\
\hline
20\% & 5 & 0.7905+-0.0003 & 75 & 0.7971+-0.0005 & 69 & 0.03 \\
\hline
50\% & 5 & 0.7903+-0.0003 & 80 & 0.7932+-0.0005 & 75 & 0.03\\
\hline
100\% & 5 & 0.7888+-0.0002 & 75 & 0.7887+-0.0003 & 72 & 0.78\\
\hline
\end{tabular}
\end{adjustwidth}
\end{table}

For all combinations of $C$s and $E$s, our method exhibited less computational complexity (that is, fewer average epochs) than $FedAvg$. With $C$=10\%, 20\% and 50\%, our method consistently achieved higher cross validation AUCs than $FedAvg$ ($p$=0.03); with $C$ =100\%, the latter’s AUC was marginally higher (0.7888 versus 0.7887, and $p$=0.78). However, implementing $C$ of 100\% might not be beneficial in practice, because involving all clients in federated learning was computationally costly and would not necessarily lead to the best predictive performance (0.7905 for FedAvg with $C$=20\% and 0.7940 for LoAdaBoost with $C$=10\%). 

\subsection*{Evaluation in non-IID scenario}
The data distribution became non-IID after sorting the examples by age and gender. FedAvg with data-sharing \cite{zhao2018federated} was the state-of-the-art method that narrowed the performance gap between IID and non-IID \cite{zhao2018federated}. The data-sharing strategy implemented on FedAvg could effectively counter the adverse effect of non-IID data distributions. To facilitate a fair comparison, we adopted the strategy and evaluated LoAdaBoost with data-sharing against Zhao \textit{et al}'s method. Like IID, we prepared data for cross validation by partitioning the non-IID examples into 90 clients, each holding 300 examples, and randomly divided the clients into 10 folds, each containing nine clients.

\begin{figure}[!h]
\centering
\includegraphics[width=0.8\linewidth]{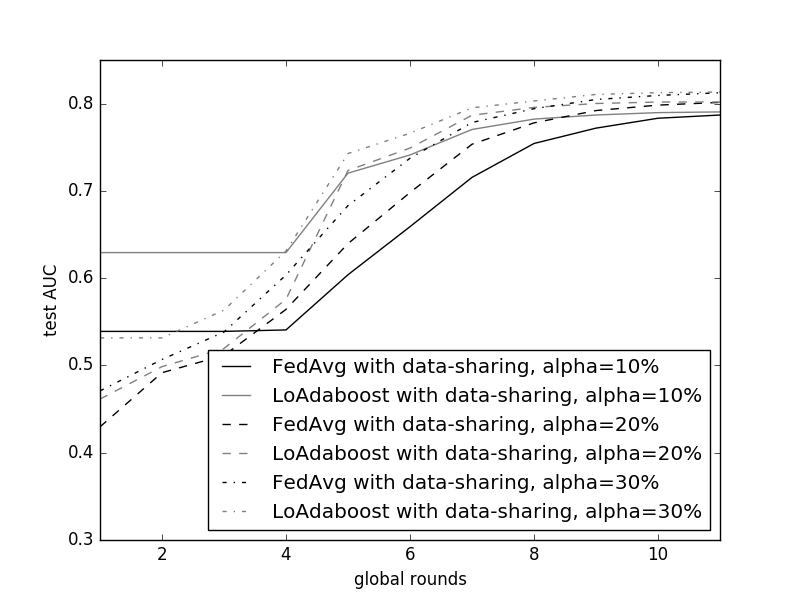}
\caption{\bf Comparison of $FedAvg$ and $LoAdaboost$ on non-IID data with data-sharing strategy.}
\label{fig:3}
\end{figure}

Figure \ref{fig:3} compares predictive performance (test AUC versus global rounds) of FedAvg and LoAdaboost with the distribution fraction $\alpha$ =10\%, 20\% and 30\%, respectively. The globally shared data size $\beta$, client fraction $C$ and epoch count $E$ were set to 1\%, 10\% and 5, respectively. For all $\alpha$s, both methods started convergence by the 10th global round; given the same $\alpha$, our method achieved a higher test AUC than $FedAvg$. 

Unlike IID evaluation where our method converged slower than $FedAvg$, here both methods had roughly the same convergence speed. We speculate the reason to be that learning on each client model with non-IID data became more difficult than with IID data, and so training for constantly five epochs across all client models was no longer advantageous. 

Same as IID evaluation, 10-fold cross validation was performed for five times. We fixed $C$ to 10\% and $E$ to 5 while varying $\alpha$ from 10\% to 30\% and $\beta$ from 1\% to 3\%. As shown in Table \ref{tab:2}, both methods’ AUCs at convergence increased with a larger value of $\alpha$ or $\beta$ (that is, more data was shared with each client). More importantly, our method always achieved a higher AUC with fewer average epochs. 

\begin{table}[!ht]
\begin{adjustwidth}{-2.25in}{0in} 
\centering
\caption{\label{tab:2}\bf Non-IID scenario: 10-fold cross validation results with varying $\alpha$ and $\beta$.}
\begin{tabular}{|P{0.7cm}|P{1.0cm}|P{3.0cm}|P{2.3cm}|P{3.0cm}|P{2.3cm}|P{1cm}|}
\hline
\multirow{2}{*}{$\beta$}&\multirow{2}{*}{$\alpha$}&\multicolumn{2}{c|}{FedAvg with data sharing}&\multicolumn{2}{c|}{LoAdaBoost with data sharing}&\multirow{2}{*}{$p$-value} \\
\cline{3-6}
& & AUC & average epochs & AUC & average epochs & \\
\hline
\multirow{3}{*}{1\%} & 10\% &	0.7842+-0.0016	& 40 & 0.7916+-0.0015 &	36	& 0.03\\
\cline{2-7}
&20\% &	0.7954+-0.0012 &	40 &	0.8016+-0.0015 &	35 &	0.03 \\
\cline{2-7}
&30\% &	0.8167+-0.0011 &	40 &	0.8203+-0.0011 &	34 &	0.03\\
\hline
2\%	& 10\% &	0.7913+-0.0010 &	40 &	0.7984+-0.0008 &	35 &	0.03 \\
\hline
3\%	& 10\% &	0.8033+-0.0010 &	40 &	0.8063+-0.0010 &	34 &	0.03 \\
\hline
\end{tabular}
\end{adjustwidth}
\end{table}

With $\alpha$=20\% and $\beta$=1\% (that is, each client received only 54 additional examples, 0.2\% of the total data), both methods obtained higher cross validation AUCs than those in IID scenario (0.7954 versus 0.7842 for $FedAvg$ with data-sharing and 0.8016 versus 0.7916 for $LoAdaBoost$ with data-sharing). Furthermore, it is worth mentioning the trade-off between the size of shared data and predictive accuracy: if more data was distributed across the clients, the higher AUCs would be obtained, and vice versa. 

Moreover, we further investigated the effect of increasing client percentage on predictive performance by fixing $\alpha$=10\%, $\beta$=1\% and $E$=5 and varying $C$. The 10-fold cross validation results are displayed in Table \ref{tab:3}. Our method obtained higher cross validation AUCs than $FedAvg$ with data-sharing with $C$=10\%, 20\%, 50\% and 100\%, and in all cases each client model under $LoAdaboost$ with data-sharing was expected to run less epochs per global round than under $FedAvg$ with data-sharing.

\begin{table}[!ht]
\begin{adjustwidth}{-2.25in}{0in} 
\centering
\caption{\label{tab:3}\bf Non-IID scenario: 10-fold cross validation results with varying $C$.}
\begin{tabular}{|P{1cm}|P{3.0cm}|P{2.3cm}|P{3.0cm}|P{2.3cm}|P{1cm}|}
\hline
\multirow{2}{*}{$C$}&\multicolumn{2}{c|}{FedAvg with data sharing}&\multicolumn{2}{c|}{LoAdaBoost with data sharing}&\multirow{2}{*}{$p$-value} \\
\cline{2-5}
& AUC & average epochs & AUC & average epochs & \\
\hline
10\%	& 0.7842+-0.0016 &	40	& 0.7916+-0.0015 &	36 &	0.03 \\
\hline
20\% &	0.7869+-0.0008 &	50 &	0.7893+-0.0005 &	46 &	0.03 \\
\hline
50\% &	0.7831+-0.0005 &	40 &	0.7877+-0.0006 &	35 &	0.03 \\
\hline
100\% &	0.7609+-0.0004 &	40 &	0.7900+-0.0003 &	35 &	0.03 \\
\hline
\end{tabular}
\end{adjustwidth}
\end{table}

\subsection*{Evaluation on eICU data}
To demonstrate the robustness of our method, we included in experiments another critical care dataset from the eICU Collaborative Research Database \cite{pollard2018eicu}. The eICU data was in nature non-IID, containing patient data from different hospitals across the US. We sampled 9,000 examples from 30 hospitals, each consisting of 300 examples and serving as a client in the non-IID scenario. The summary of this data is shown in Table \ref{tab:4}. 

\begin{table}[!ht]
\begin{adjustwidth}{-2.25in}{0in} 
\centering
\caption{\label{tab:4}\bf Summary of the eICU dataset.}
\begin{tabular}{ |c|c|c| }
 \hline
 \empty & representation &  count\\
 \hline
$PATIENT\_UNIT\_STAY\_ID$ & integer: six-digit patient ID & 22,500 \\
 \hline
$HOSPITAL\_ID$	& integer: hospital IDs ranging from 63 to 458	&45 \\
 \hline
$MORTALITY$	& binary: 0 for survival and 1 for expired	& 21393/1107 \\
 \hline
$DRUGS$	& binary: 0 for not prescribed to patients and 1 for prescribed	& 1399 dimensions \\
 \hline
\end{tabular}
\end{adjustwidth}
\end{table}

\begin{figure}[!h]
\centering
\includegraphics[width=0.8\linewidth]{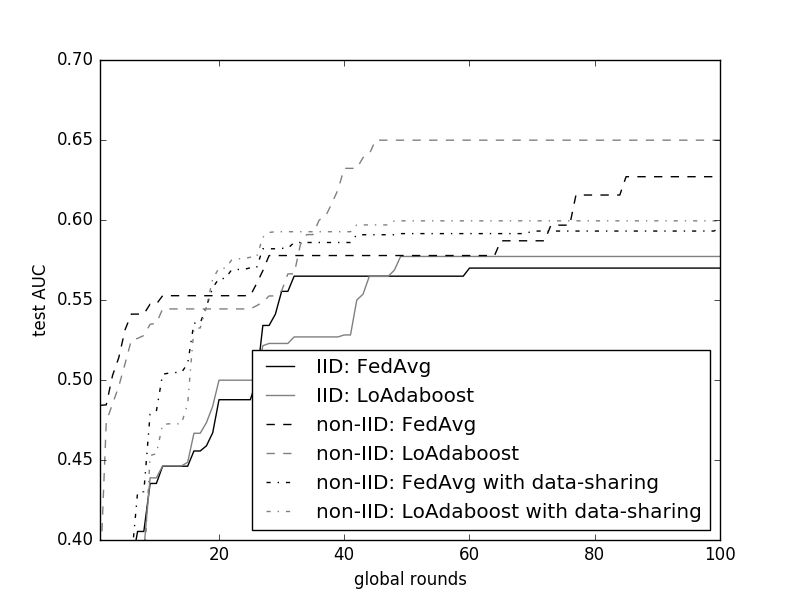}
\caption{\bf Comparison of $FedAvg$ and $LoAdaboost FedAvg$ on eICU data.}
\label{fig:4}
\end{figure}

Same as MIMIC III, DRUGS prescribed to patients during the first 48 hours of stay were used to predict MORTALITY of patients. In addition, another randomly chosen 90 examples was prepared as the holdout set (that is, $\beta$=1\%) for implementing the data-sharing strategy. For IID evaluation, we shuffled those 9,000 examples and then partitioned them into 30 clients, each containing 300 examples. The clients were randomly divided into 10 equally-sized folds. Nine folds were regarded as the training set and the remaining fold was used as the test set. Throughout the evaluations, $C$ and $E$ were set to 10\% and 5, respectively. In non-IID scenario with data-sharing strategy, $\alpha$ was set to 10\%. Figure \ref{fig:4} shows the evaluation results of a single run of cross validation. 

Federated learning outcomes on eICU were different from those on MIMIC III data. Learning became more difficult as both the baseline and our method took 50 or more global rounds to converge. In addition, as displayed in the figure, AUCs with non-IID data were close to 0.65 but dropped to roughly 0.6 when data-sharing was adopted, while AUCs with IID data were notably lower for both methods. Therefore, learning on non-IID seemed easier than on IID, which resonated with the evaluation results of language modeling on the Shakespeare dataset in McMahan \textit{et al}.’s work \cite{mcmahan2016communication}. What was consistent with evaluation on MIMIC III data was that $LoAdaBoost$ converged to higher AUCs with fewer average epochs than $FedAvg$, whether the scenario be IID, non-IID or non-IID with data-sharing. This finding was confirmed by the results of 10-fold cross validation with five repetitions (see Table \ref{tab:5}). 

\begin{table}[!ht]
\begin{adjustwidth}{-2.25in}{0in} 
\centering
\caption{\label{tab:5}\bf Evaluation on eICU data: 10-fold cross validation results.}
\begin{tabular}{|P{3.5cm}|P{5.0cm}|P{2.5cm}|P{2.3cm}|P{1.2cm}|}
\hline
data distribution & method & AUC & average epochs & $p$-value \\
\hline
\multirow{2}{*}{IID} & FedAvg & 0.5693+-0.0057 &400&\multirow{2}{*}{0.03} \\
\cline{2-4}
& LoAdaBoost &0.6057+-0.0077& 262 & \\
\hline
\multirow{4}{*}{non-IID} &FedAvg& 0.6512+-0.0043& 300& \multirow{2}{*}{0.03}\\
\cline{2-4}
 &LoAdaBoost& 0.6548+-0.0048& 271& \\
\cline{2-5}
&FedAvg with data-sharing& 0.6253+-0.0088& 350& \multirow{2}{*}{0.03}\\
\cline{2-4}
 &LoAdaBoost with data-sharing& 0.6412+-0.0065& 272& \\
\hline
\end{tabular}
\end{adjustwidth}
\end{table}

\section*{Discussion}
Distributed health data in large quantity and of high privacy can be harnessed by federated learning where both data and computation are kept on the clients. In this study, we proposed \textit{LoAdaBoost FedAvg} that adaptively boosted the performance of individual clients according to cross-entropy loss. Under the federated learning scheme, the data held on each client was random in IID scenario and came from different distributions in non-IID scenario; and the randomly chosen clients participating in each round of learning would also be different. Therefore, if the number of epochs $E$ was fixed as in the case of FedAvg, there could highly likely be certain underfitted or overfitted clients at each global round, which would adversely affect model averaging at the server. On the other hand, our method firstly trained each client for very few epochs, then defined the goodness-of-fit of each client by comparing its cross-entropy loss with the median loss from the previous round, and finally achieved performance boosting by further training poorly-fitted clients for more epochs, well-fitted ones for less, and over-fitted ones for none. In this manner, all clients would expectedly be more appropriately learnt than those of FedAvg. Experimental results with IID data and non-IID data showed that \textit{LoAdaBoost FedAvg} converged to slighly higher AUCs and consumed fewer average epochs of clients than $FedAvg$. Our approach can also be extended to learning tasks in other fields, such as image classification and speech recognition, wherever the data is distributed.

As a final point, federated learning with IID data does not always outperform that with non-IID data. Evaluation on the eICU data is such an example; and another one is the language modeling task on the Shakespeare dataset\cite{mcmahan2016communication} where learning on the non-IID distribution reached the target test-set AUC nearly six times faster than on IID. In cases like this, the data-sharing strategy becomes unnecessary. Moreover, according to Zhao \textit{et al.}\cite{zhao2018federated}, weight divergence would occur in neural network models trained on clients holding data from different distributions, and was positively correlated with the degree of data skewness. The predictive accuracy of $FedAvg$ could be reduced by up to 55\% due to high weight divergence. When non-IID data is severely skewed, \textit{LoAdaBoost} may also lose its competitive advantage. This is because the weights of clients' models can all diverge from the well-tuned weight that could have been obtained in centralized learning\cite{zhao2018federated}, and the measure of median client-training loss may no longer be an effective indicator of the overall training quality of federated learning. In the continuation of our study, we will investigate what kind of medical datasets may result in superior modeling performance with non-IID distribution and why this occurs. Furthermore, we will try to improve the LoAdaBoost FedAvg algorithm to make learning on such datasets even easier.


\section*{Author contributions statement}

L.H initiated the idea, designed the algorithm, processed the data and conducted the experiments. Y.Y conducted the experiments and processed the data. Z.F conducted the experiments and help design the algorithm. S.Z instructed computational optimization and realization. H.D is a clinical expert experienced critical care and provided clinical instructions in this project. D.L initiated the idea, designed the algorithms, supervised and coordinated the project. All authors reviewed the manuscript. 

\section*{Additional information}
Competing Interests: The authors declare no competing interests.


%
%
%

\end{document}